%% file: main.tex
\documentclass[letterpaper, 10 pt, conference]{IEEEtran}  % Comment this line out
\input{tex/preamble}
\usepackage{graphics} % for pdf, bitmapped graphics files
\usepackage{graphicx}
\usepackage{epsfig} % for postscript graphics files
\usepackage{times} % assumes new font selection scheme installed
\usepackage{physics}
\usepackage{xcolor}
\usepackage{tikz}

\usetikzlibrary{calc} 
\usetikzlibrary{positioning}
\usepackage{balance}
\usepackage{soul}
\usetikzlibrary{arrows}
\usepackage{amsmath}
\usepackage{diagbox}
\usepackage{hyperref}
\usepackage[top=2.54cm, bottom=1.91cm, left=1.91cm, right=1.91cm]{geometry}

\IEEEoverridecommandlockouts
\title{\LARGE \bf%
TacDiffusion: Force-domain Diffusion Policy for Precise Tactile Manipulation}

\author{Yansong Wu\textsuperscript{1*}, Zongxie Chen\textsuperscript{1*}, Fan Wu$^{1\dagger}$, Lingyun Chen$^{1,2}$, Liding Zhang$^{1}$, \\Zhenshan Bing$^{1, 4}$, Abdalla Swikir$^{1,3}$, Sami Haddadin$^{1,3}$, Alois Knoll$^{1}$ % <-this % stops a space
\thanks{
\textsuperscript{*} Equal contribution. 

$^\dagger$ Corresponding author: Fan Wu ({\tt\small f.wu@tum.de}). 

$^{1}$ Munich Institute of Robotics and Machine Intelligence (MIRMI), Technical University of Munich, Germany. 
$^{2}$ Centre for Tactile Internet with Human-in-the-Loop (CeTI).
$^{3}$ Mohamed Bin Zayed University of Artificial Intelligence, Abu Dhabi, UAE.
$^{4}$ State Key Laboratory for Novel Software Technology and the School of Science and Technology, Nanjing University (Suzhou Campus), China.

Code available:~\href{https://github.com/popnut123/TacDiffusion}{https://github.com/popnut123/TacDiffusion}
}%

}

\renewcommand{\baselinestretch}{0.905}

\let\oldtwocolumn\twocolumn
\renewcommand\twocolumn[1][]{%
    \oldtwocolumn[{#1}{
    \vspace{-20pt}
    \begin{center}
           \includegraphics[width=\textwidth]{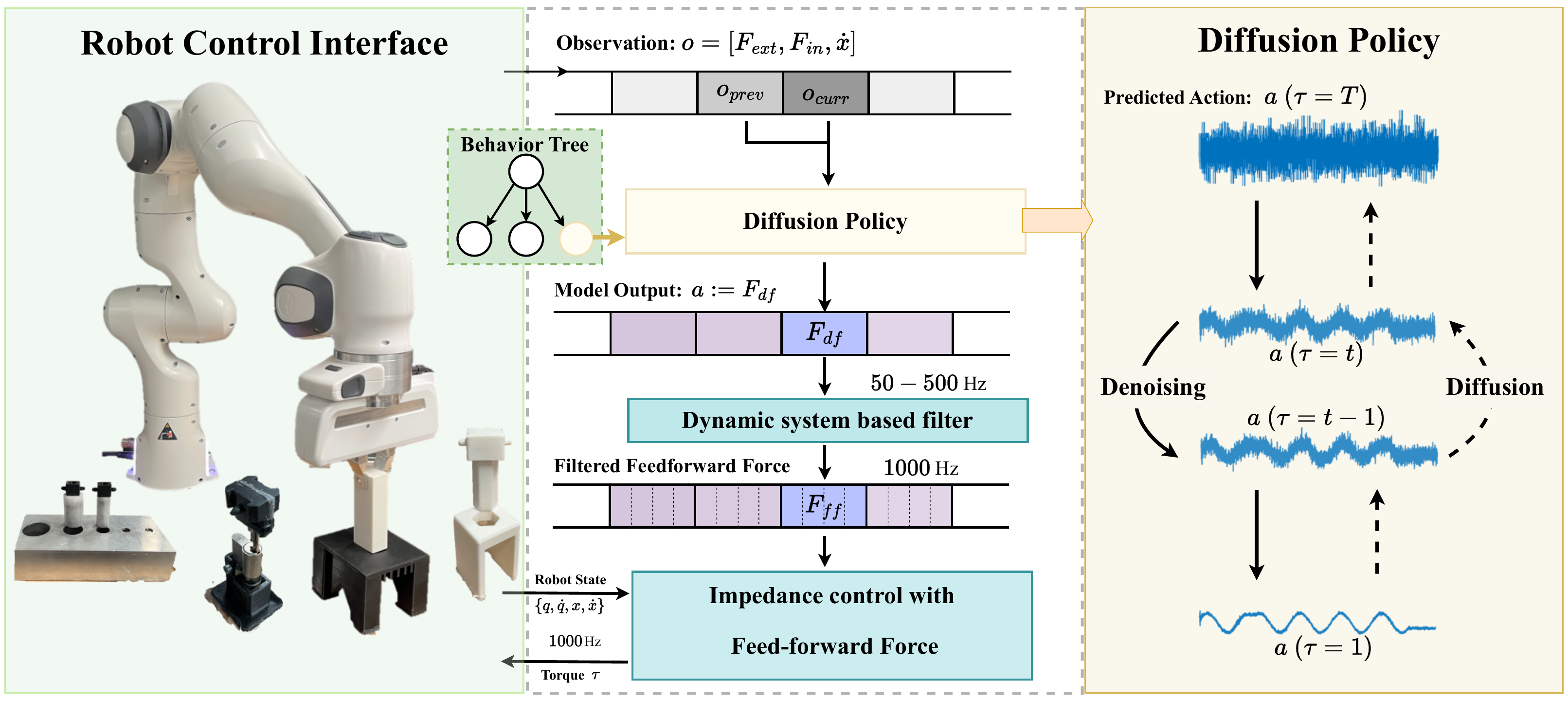}
           \captionof{figure}{Framework Overview: \textit{TacDiffusion} processes observations~$\bm{o}$, including both tactile and spatial information, to generate 6D wrench~$\bm{F}_{df}$. It addresses the frequency misalignment between the diffusion model and the real-time controller through a dynamic system-based filter, and regulates robot’s force and motion behavior using impedance control with feed-forward force~$\bm{F}_{ff}$.}
           \label{fig:IL_overview_new}
    \end{center}
    \vspace{5pt}
    }]
}

\begin{document}
\maketitle
% \usage %comment this to remove usage page
%\input{tex/usage}

% \begin{strip}\centering
% \includegraphics[width=\textwidth]{figures/Overall.pdf}
% \captionof{figure}{Framework Overview. Tactile information, spatial information.
% \label{fig:IL_overview_new}}
% \end{strip}

% \maketitle

% \twocolumn[{
% \maketitle
% \begin{center}
%     \captionsetup{type=figure}
%     \includegraphics[width=0.98\textwidth]{figures/Overall.pdf}
%     \captionof{figure}{Framework Overview. Tactile information, spatial information.
% \label{fig:IL_overview_new}}
% \end{center}
% }]

% \setcounter{figure}{1}
% \makeatletter
% \let\@oldmaketitle\@maketitle%
% \renewcommand{\@maketitle}{
%    \@oldmaketitle%
%    \begin{center}
%     \centering
%     \includegraphics[width=0.98\textwidth, clip]
%     {figures/Overall.pdf}
%   \end{center}
%   \footnotesize{\textbf{Fig.~\thefigure:\label{fig:IL_overview_new}}~ Framework Overview. Tactile information, spatial information 
%   }
% }
% \makeatother
% \maketitle

%\thispagestyle{empty}
%\pagestyle{empty}

\begin{abstract}

% Assembly is an essential skill for robots in both modern manufacturing and service robotics. Achieving high-precision in robotic assembly tasks, particularly in dynamic environments, remains a significant challenge. This paper introduces \textit{TacDiffusion}, a pioneering framework that leverages diffusion models to generate continuous 6D wrench actions for tactile manipulation. Unlike traditional methods relying on impedance control, \textit{TacDiffusion} enhances both task-level transferability and control-level adaptability. Our approach, the first to apply diffusion models in the force-domain, shows remarkable improvements in zero-shot transferability, achieving a 95.7\% success rate across diverse high-precision tasks. We also address the trade-off between model accuracy and inference speed by providing practical guidelines for model selection and introducing a dynamic filter to resolve frequency misalignment between the diffusion model and real-time control, resulting in a 9.15\% increase in task success rates. Extensive experiments validate the framework's effectiveness in real-world settings, offering a robust solution for high-precision tactile manipulation.

% Assembly is an essential skill for robots in both modern manufacturing and service robotics. However, learning transferable insertion skill capable of handling diverse high-precision assembly tasks remains a significant challenge. This paper introduces the first framework that leverages diffusion models to generate 6D wrench for high-precision tactile robotic insertion tasks. 

Assembly is a crucial skill for robots in both modern manufacturing and service robotics. However, mastering transferable insertion skills that can handle a variety of high-precision assembly tasks remains a significant challenge. This paper presents a novel framework that utilizes diffusion models to generate 6D wrench for high-precision tactile robotic insertion tasks. It learns from demonstrations performed on a single task and achieves a zero-shot transfer success rate of 95.7\% across various novel high-precision tasks. Our method effectively inherits the self-adaptability demonstrated by our previous work. In this framework, we address the frequency misalignment between the diffusion policy and the real-time control loop with a dynamic system-based filter, significantly improving the task success rate by 9.15\%. Furthermore, we provide a practical guideline regarding the trade-off between diffusion models' inference ability and speed.

% analyze the frequency misalignment between the diffusion model and force-domain actions, providing a systematic guideline for selecting the appropriate model size to balance the model's inference ability and speed. Additionally, we design a dynamic system-based filter that effectively mitigates the conflict between inference ability and speed, showcasing a 9.15\% improvement in the success rate.

% Assembly is an essential skill for robots in both modern manufacturing and services robotics. However, achieving broad adaptability across diverse high-precision assembly tasks remains a significant challenge. To address this problem, this paper introduces the first framework that leverages diffusion models to generate force-domain actions for robotic insertion tasks, achieving high data-efficiency and broad adaptivity. Our system learns from demonstrations of a single task and achieves a 95.7\% zero-shot transfer success rate on diverse novel tasks. In this framework, we highlight the frequency misalignment issue between the diffusion model and force-domain actions and provide a systematic guideline to choose the proper model size to balance the inference ability and model size. Furthermore, we design a dynamic system based filter which effectively mitigates this issue.

% Keywords: Robotics, Diffusion Model, High-precision assembly, Zero-shot Transfer, Data-efficiency Learning, 
\end{abstract}

\input{intro}
\input{related_works}

\input{2_methods}

\input{3_experiment}
\input{4_conclusion}

\section*{ACKNOWLEDGMENT}
The authors acknowledge the financial support by the Bavarian State Ministry for Economic Affairs, Regional Development and Energy (StMWi) for the Lighthouse Initiative KI.FABRIK (Phase 1: Infrastructure as well as the research and development program under, grant no. DIK0249), the European Union’s Horizon 2020 research and innovation programme as part of the project euROBIN under grant no. 10107059, the Federal Ministry of Education and Research of Germany (BMBF) in the programme of ``Souverän. Digital. Vernetzt." Joint project 6G-life, project identification number 16KISK002, and by the German Research Foundation (DFG, Deutsche Forschungsgemeinschaft) as part of Germany’s Excellence Strategy – EXC 2050/1 – Project ID 390696704 – Cluster of Excellence “Centre for Tactile Internet with Human-in-the-Loop” (CeTI) of Technische Universität Dresden.

\bibliography{IEEEabrv,mybib2022}
\bibliographystyle{myIEEEtran}

%\appendix

\label{last-page}
\end{document}

%% file: tex/preamble.tex
\let\labelindent\relax         % prevents a conflict with IEEE \labelindent when using enumitem

\usepackage{times}
\usepackage{graphicx}%\graphicspath{{figures/}} 
\usepackage{epstopdf}
\usepackage{mdwmath}
\usepackage{mdwtab}
\usepackage{rotating}
\usepackage{caption}
\usepackage{subcaption}

\usepackage{amssymb}
\usepackage{amsfonts}
\usepackage{amsmath}
\usepackage{amsthm}
\usepackage{bm}

\usepackage{algorithm}
\usepackage{cite}
\usepackage{algorithmic}

\usepackage{multirow} 
\usepackage{multicol}
\usepackage{array}

\usepackage{standalone}
\usepackage{booktabs} % For \toprule, \midrule and \bottomrule

\usepackage{siunitx} % Formats the units and values

\usepackage{accents}

\usepackage{pgfplotstable} % Generates table from csv
\usepackage{verbatim}
\usepackage{isomath}
\usepackage{overpic}

\usepackage{tikz}
\usetikzlibrary{shapes,calc,patterns,
	decorations.pathmorphing,
	decorations.markings}

\tikzstyle{spring}=[very thick,decorate,decoration={zigzag,pre length=2,post
	length=2,segment length=6}]

\tikzstyle{damper}=[thick,decoration={markings, 
	mark connection node=dmp,
	mark=at position 0.5 with 
	{
		\node (dmp) [thick,inner sep=0pt,transform shape,rotate=-90,minimum
		width=15pt,minimum height=3pt,draw=none] {};
		\draw [thick] ( $(dmp.north east)+(2pt,0)$ ) -- (dmp.south east) -- (dmp.south
		west) -- ( $(dmp.north west)+(2pt,0)$ );
		\draw [thick] ( $(dmp.north)+(0,-5pt)$ ) -- ( $(dmp.north)+(0,5pt)$ );
	}
}, decorate]

\makeatletter\newcommand{\manuallabel}[2]{\def\@currentlabel{#2}\label{#1}}\makeatother

\input{tex/colours}       % colour definitions
\input{tex/symbols}       % symbol definitions
\input{tex/graphics}      % graphics definitions
\input{tex/format}        % formatting definitions
\input{tex/comments}      % set up commenting

%% file: tex/colours.tex
\usepackage{color}
\usepackage{xcolor}

\colorlet   {lightorange}{orange!20}
\colorlet   {lightgrey}  {gray!20}

%% file: tex/symbols.tex
\usepackage{amssymb}
\usepackage{mathtools}

\mathchardef\mhyphen="2D   % define "math hyphen"

\newcommand{\RNum}[1]{\uppercase\expandafter{\romannumeral #1\relax}}

\newcommand{\be}{\bm{e}} %{\mathbf{e}}
\newcommand{\bq}{\bm{q}}     %{\mathbf{q}}         % (bold) joint angles
\newcommand{\bx}{\bm{x}}     %{\mathbf{x}}         % (bold) state
\newcommand{\btau}{\boldsymbol{\tau}}

%% file: tex/graphics.tex
\graphicspath{
{figures/}
}

%% file: tex/format.tex
\providecommand{\figurename}{Fig.}
\usepackage[inline]{enumitem}
\setlist{nolistsep}
\newcommand{\il}[1]{\begin{enumerate*}[label=(\roman*)]#1\end{enumerate*}}

%% file: tex/comments.tex
\usepackage[normalem]{ulem}                                        % <- for striking text out
\usepackage{marginnote}
\usepackage[textwidth=10ex,colorinlistoftodos]{todonotes}
\colorlet{fwu}{red}
\colorlet{ywu}{blue}
\colorlet{zbing}{green}

%% file: intro.tex
\section{Introduction}
Assembly tasks are crucial in robotics, serving as the backbone of modern manufacturing and service applications~\cite{whitney2004mechanical}. 
As the demand for flexible manufacturing grows, robotic assembly increasingly takes place in dynamic environments, where objects are not precisely positioned at known locations and part holders are often not viable~\cite{Nottensteiner2021Autonomousa}.
Achieving both broad transferability and precise control capabilities in these conditions remains a significant challenge.
Human workers, on the other hand, demonstrate exceptional dexterity in assembling diverse objects with tight-clearance components, primarily by leveraging tactile feedback from their fingertips throughout the process~\cite{johansson1983tactile, birznieks2001encoding}. Similarly, a versatile high-precision robotic assembly system must exhibit both task-level transferability---generalizing across a wide range of objects and parts---and control-level self-adaptability, enabling it to respond to environmental changes often sensed through tactile feedback~\cite{li2019survey, nottensteiner2021towards}.

%what to write for 2nd paragraph:
%- tactile feedback and force control is important for assembly.
%- but, due to several challenges, the robot learning community favor simpler action space using position and velocity, and control force indirectly via impedance control. 
%- but, impedance control is insufficient, dexterous manipulation often require regulating motion and force simultaneously.
%- despite recent successes in transformer-based frameworks and diffusion models as policy representation, most prior works rarely address precise force control. 
Throughout the history of robotics research, the importance of tactile feedback and force control for high-precision assembly has been consistently recognized~\cite{Inoue1974Force,chen2007integrated, chen2009high, inoue2017deep, johannsmeier2019framework, luo2019reinforcement, li2019survey, beltran2020variable}. 
However, several challenges persist in precise force control, including the difficulty of accessing to appropriate robot hardware and expensive force sensors, %the increased computational complexity for learning due to higher dimensionality, 
the complexity of ensuring stability and safety while regulating force, the sensitivity of force control to environmental changes, the difficulty of estimating environment constraints and contact dynamics in dynamic settings, and the challenge of collecting high-quality tactile data for learning force control.
Due to these barriers, the use of simpler motion-domain action spaces, with impedance control as an indirect force control method is often favored by the robot learning community. 
Nevertheless, the increasing diversity of contact-rich manipulation tasks highlights the equal importance of simultaneously regulating motion, compliance, and force, so that agents can autonomously perform a wide range of task stably and robustly, without the need for explicit controller switching \cite{Haddadin2024Unified}. 
Despite the recent successes of implementing transformer~\cite{driess2023palm, rt1, rt2, shridhar2023perceiver, rtx, Goyal2024RVT2} and/or diffusion-based~\cite{pearce2022imitating,carvalho2023motion,kapelyukh2023dall, mishra2023generative, black2023zero, chi2023diffusion, reuss2023goal, mishra2024reorientdiff, sridhar2024nomad, li2024generalization, meng2025preserving} policies for robot manipulation that exhibit excellent generalization capability, 
it remains unexplored how to integrate force control with these models for high-precision tactile manipulation, so that the benefits of these generative models for multi-modal modelling and prediction can be fully exploited.

To address this gap, and aiming to achieve both task-level \emph{transferability} and control-level \emph{self-adaptability}, we propose \textit{TacDiffusion}, a novel framework that leverages a diffusion policy for high-precision tactile manipulation. To the authors' knowledge, it is the first framework to employ diffusion models in generating force-domain actions for tactile-based robotic manipulation in tight-clearance insertion tasks. \textit{TacDiffusion} learns from demonstrations performed by expert policies on a single task and achieves an overall 95.7\% zero-shot transfer success rate across various novel high-precision, \textit{sub-millimeter-level} peg-in-hole tasks. By imitating the expert policies, which are based on a behavior tree-based skill proposed in our previous work~\cite{10610835}, \textit{TacDiffusion} successfully inherits its self-adaptability, characterized by the ability to switch skill primitives based on real-time tactile sensing. Importantly, compared to the expert policy, \textit{TacDiffusion} also outperforms in execution time and robustness on these novel tasks in a zero-shot transfer manner. 

To further enhance real-time performance, we investigate how model size affects the trade-off between accuracy and inference speed, providing practical guidelines for optimal model selection. Moreover, to handle the frequency misalignment between the diffusion policy's inference process and the low-level controller, a dynamic system-based filter is designed to smooth the output of the diffusion model for high-frequency force-impedance control, significantly improving the task success rate by 9.15\%.

In summary, our main contributions are:
\il{
    \item a novel diffusion-based policy that outputs 6D wrench for tactile manipulation;
    \item learning from a behavior tree-based expert policy to inherent its tactile-based self-adaptability;
    \item a dynamic system-based filter smoothing and aligning low frequency outputs from diffusion model with high frequency control, with experimental evidences showing significant effect on task performance;
    \item investigation on trade-off between accuracy and inference speed, resulting in insights for optimal model selection in practice.
}

%% file: related_works.tex
\section{Related Works}
In this section, we focus our review on \il{\item High-Precision Assembly Tasks, \item Transferability and \item Diffusion model in robotics.}

\subsection{High-Precision Assembly Tasks}
Due to the robot's accuracy limitation, position-based control methods are insufficient for high-precision assembly tasks that require accuracy exceeding the robot's precision \cite{inoue2017deep}. To address this issue, recent studies have shifted to designing actions in the force-domain rather than the position domain to perform high-precision robotic assembly tasks. According to the control strategy, these methods span four main categories: force controller \cite{chen2009high}, admittance controller \cite{chen2007integrated}, hybrid position/force controller\cite{inoue2017deep, beltran2020variable}, and impedance controller with feed-forward force \cite{johannsmeier2019framework, luo2019reinforcement}. Nevertheless, these works normally focus on a specific tight-clearance task and lack investigation into the method's transferability and adaptability to novel tasks\cite{li2019survey}. 

\subsection{Enhancing Transferability in Robotic Assembly}
In the last decade, there is extensive literature on generating robotic assembly policies with broad generalization. Deep Reinforcement Learning-based methods, for instance, typically achieve the generalization ability through training with multiple objects \cite{spector2021insertionnet, spector2022insertionnet}. Another noteworthy case is meta-learning, which trains a pre-trained model using online or offline data from a diverse and comprehensive set of tasks, enabling domain adaptation ability through fine-tuning \cite{schoettler2020meta, zhao2022offline, bing2023meta}. Furthermore, sim-to-real based approaches have gained attention for their cost-effective data collection in the simulation environments, and zero-shot sim-to-real transfer for perception-initialized assembly has been only recently demonstrated \cite{tang2023industreal, tang2024automate}. Besides, to tackle precise manipulation, RVT-2 \cite{Goyal2024RVT2} trained a transformer-based multi-task policy. Despite improving performance on multi-task learning benchmark, its success rate on high-precision (millimeter level) insertion tasks, roughly 50\%, is far from being satisfactory to deploy to real assembly production. 
Aside from these approaches, evolutionary algorithms with parameterized robot skills have shown transferability across tasks via fine-tuning \cite{10610835}. However, achieving zero-shot transfer on high-precision tasks with a satisfactory success rate in the real world remains an open challenge
% All these methods share the same goal: improving data-efficiency, specifically by using less training data, fewer training objects, and fewer fine-tuning steps while enhancing transferability to as many novel objects as possible.

% In contrast to the traditional discriminative models, the diffusion models capture the entire data distribution rather than focusing solely on decision boundaries.

% \cite{schoettler2020meta} meta - We adapt the same policy, which was learned in simulation, to each of the two tasks, despite their distinct physical properties. Moreover, in each task, our method adapts with just 20 trials, significantly fewer than in previous work. 
% \cite{zhao2022offline} meta - offline 30 minutes of adaptation
% \cite{10610835}
% \cite{tang2024automate} automate: zero-shot sim-to-real transfer
% \cite{spector2021insertionnet, spector2022insertionnet} Insertion net, training with multiple objects
% \cite{tang2023industreal} IndustrialReal: We present IndustReal, a set of algorithms, systems, and
% tools for solving contact-rich assembly tasks in simulation and transferring behaviors to reality

\subsection{Diffusion Model in Robotics} 
Meanwhile, in other areas of robotics, diffusion models \cite{ho2020denoising} have made significant progress. Compared to traditional discriminative models, diffusion models excel in generalization, achieving superior performance on unseen tasks and scenarios, by establishing a stochastic transport map between an empirically observed target distribution and a known prior \cite{li2024generalization}. Recent works have typically used scene images as input to solve planning problems \cite{kapelyukh2023dall, mishra2023generative, sridhar2024nomad} and perform manipulation tasks \cite{black2023zero, chi2023diffusion, reuss2023goal, mishra2024reorientdiff} in robotics. However, the application of diffusion models with other input modalities remains relatively underexplored in robotics, with only few studies addressing this area \cite{carvalho2023motion}. In addition, considering diffusion model applications in sequential behavior imitation \cite{pearce2022imitating} and time series processing \cite{kollovieh2024predict}, there is great potential for adapting diffusion models to force-domain actions in robotics.

% Nevertheless, the application of diffusion models with other input modalities remains underexplored in robotics. The feasibility of solving motion planning problems using high-dimensional trajectory inputs has been demonstrated \cite{carvalho2023motion}. Inspired by diffusion model applications in sequential behavior imitation \cite{pearce2022imitating} and time series processing \cite{kollovieh2024predict}, there is great potential for adapting diffusion models to time series in robotics. 

% \cite{kapelyukh2023dall} DALL-E-Bot: image input and rearrange scene image output
% \cite{carvalho2023motion} Motion planning diffusion: learn the prior model from expert data and incorporate it into optimization-based motion planning (a trajectory with high dimension as input)
% \cite{chi2023diffusion} visuiomotor: Image Observation Sequence 2 action sequence
% \cite{black2023zero} SuSIE: leveraging pretrained image-editing models to enable generalizable robotic manipulation.
% \cite{mishra2024reorientdiff} Reorient object
% \cite{mishra2023generative} Long horizon planning image input
% \cite{pearce2022imitating}  adapt diffusion models to sequential environments to mimic human behavior. 
% \cite{sridhar2024nomad} visual navigation
% \cite{reuss2023goal} manipulation: goal conditioned imitation learning

% \cite{kollovieh2024predict} -- diffusion model in time series forecasting \\

In summary, although significant progress has been made in insertion tasks, achieving zero-shot transfer in high-precision assembly tasks remains an ongoing challenge. Additionally, the application of diffusion models to force-domain actions has not yet been explored. To bridge these gaps, we propose a novel framework that leverages diffusion models to enable more efficient zero-shot transfer in high-precision insertion tasks.

%% file: 2_methods.tex
\section{Methods}
% To solve the aforementioned issues, we developed a framework with replace the ``Insertion'' subtree in our previous work~\cite{10610835} with a force-domain diffusin policy.
To solve the aforementioned issues, we develop a framework that adapts the diffusion model to force-domain actions for high-precision tactile assembly tasks. In the following subsections, we first provide an overview of the framework, followed by a detailed explanation of the concrete modules, \emph{i.e.,} the diffusion model, the impedance control with feed-forward force, and the dynamic system-based filter. 

%replaces the ``Insertion'' subtree in our previous work with a force-domain diffusion policy. The original sbu

\subsection{Framework Overview \label{sec: framework}}

Our framework comprises two key functional modules: the diffusion policy-based action generation module and the impedance control with feed-forward-based execution module.\footnote{A practical consideration here is the compatibility issues between the real-time kernel and the NVIDIA CUDA Toolkit.} As illustrated in Fig.~\ref{fig:IL_overview_new}, 
the diffusion-based policy is integrated into the behavior tree (BT) based Insertion skill by replacing the original sub-tree, which contained two primitives and a state estimator. The resultant behavior tree is simplified into a sequence of skill primitives, with ``approach" and ``contact" as two preceding primitives. As the BT is simplified into a sequence and the diffusion model handles primitive-switching, the discussion of the preceding skill primitives for contact initialization is beyond the scope of this work. For more details, we refer readers to our previous work~\cite{10610835}.
% As the BT is simplified to a sequence, and the primitive-switching behavior is captured by the diffusion model, discussion about the preceding skill primitives for contact initialization is not the focus of this work, and we refer the readers to our previous work~\cite{10610835} for a more detailed explanation. 

% the diffusion-based policy is employed within an Insertion skill represented by a behavior tree (BT) to replace the original sub-tree containing two primitives and a state estimator. 

During the assembly process, the interaction between the robot and the environment is captured as observation $\bm{o}$, which includes the external wrench, internal wrench, and end-effector's speed. The diffusion model then predicts the force-domain actions ($\bm{a}:= \bm{F}_{df}$) based on both the current observation $\bm{o}_{curr}$ and the previous observation $\bm{o}_{prev}$. Due to the restrictions of computational resources, the diffusion model's inference frequency typically ranges from 50 Hz to 500 Hz (Table~\ref{tab:Diffusion Model Parameter}), which is misaligned with the robot's 1000 Hz real-time control loop. To mitigate this, we design a dynamic system-based filter to interpolate the diffusion model’s predictions $\bm{F}_{df}$. The filtered action is then transmitted to the impedance controller with feed-forward force. Based on the desired goal $\bm{x}_d$ (insertion hole's pose) and the force command, it regulates the robot's motion and force behavior simultaneously.

\subsection{Diffusion Model \label{sec:DDPM}}

Denoising diffusion probabilistic model (DDPM)~\cite{ho2020denoising, nichol2021improved, yang2024survey} is a specific type of diffusion model designed to generate data by learning to reverse a noise injection process. DDPM consists of two processes: diffusion and denoising. The diffusion process systematically transforms the data into noise, while the denoising process is responsible for converting this noise back into data.

\begin{figure}[h]
    \centering
    \includegraphics[width=0.95\linewidth]{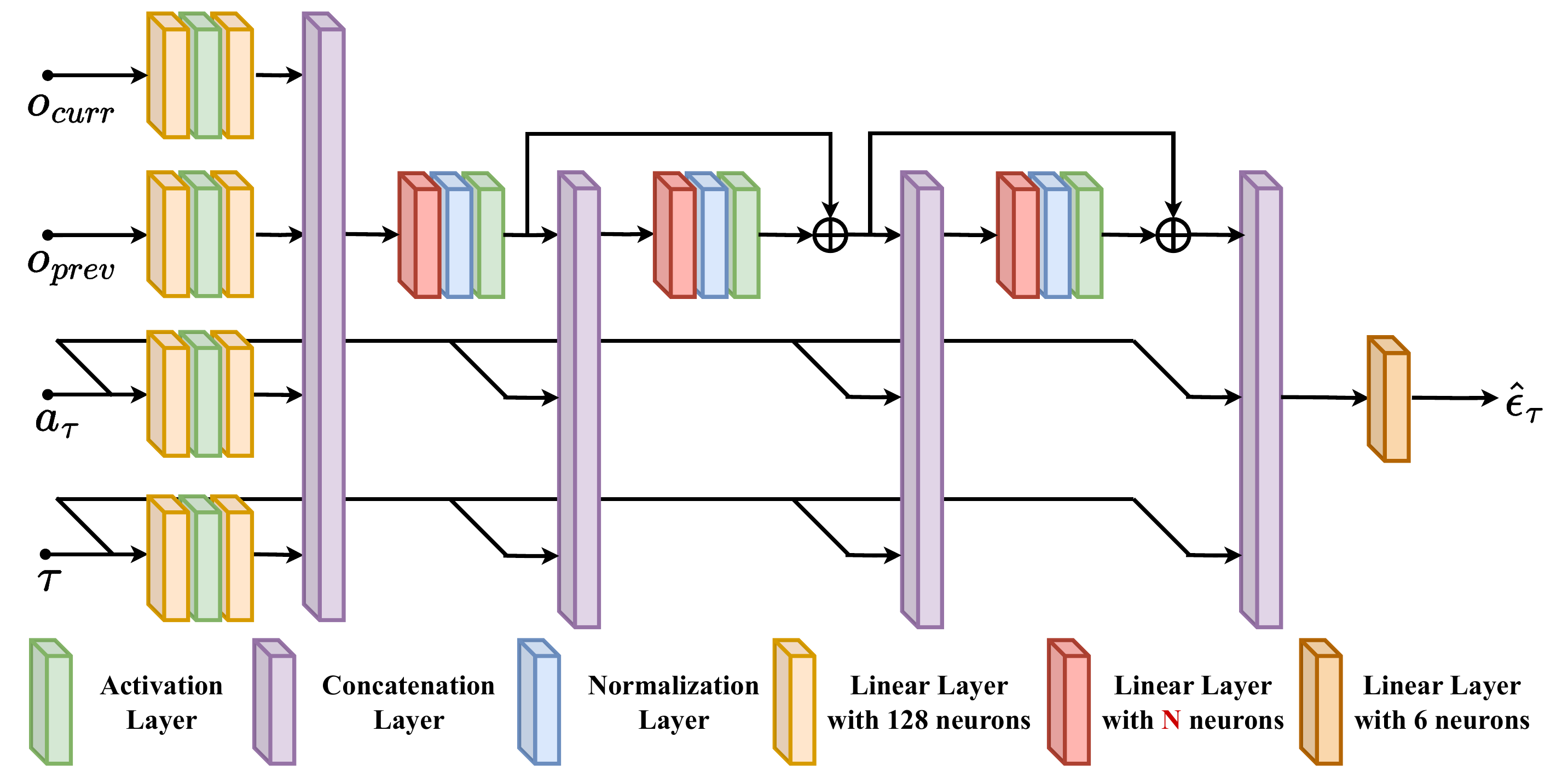}
    \caption{Network architecture of the noise estimator.}
    \label{fig:noise estimator flowchart}
\end{figure}

\subsubsection{Diffusion Process\label{sec:DDPM_diffusion}}
The \textit{diffusion} process is a forward progressive process that destructs data with noise over a series of steps. By progressively injecting noise into a ``clean" initial action $\bm{a}_{0}$, a sequence of ``polluted" actions $\bm{a}_{1}, \bm{a}_{2}, \cdots, \bm{a}_{T}$ converging to a Gaussian distribution is obtained, according to the diffusion rule~\cite{pearce2022imitating}:
\begin{align}
    {\alpha}_{\tau} &= 1 - {\beta}_{\tau} \label{eq:diffusion_law_alpha},  \\
    \bm{a}_{\tau} &= \sqrt{\alpha_{\tau}} \ \bm{a}_{\tau - 1} + \sqrt{\beta_{\tau}} \ \bm{\epsilon}_{\tau}, \label{eq:diffusion_law_main} 
\end{align}
where $\tau \in [1, T]$ denotes the diffusion step, with $T$ referring to the total number of denoising steps (not to be confused with the environment time step, as it is common in time serials). $\bm{a}_{\tau}$ and $\bm{\epsilon}_{\tau} \in \mathcal{N}(\bm{0}, \bm{I})$ represent the diffused action and the corresponding noise in the $\tau$-th diffusion step. ${\alpha}_{\tau}$ and ${\beta}_{\tau}$ refer to variance schedule parameters that regulate the noise mixed in each diffusion step.

Furthermore, the noise $\bm{\epsilon}_\tau$ also plays a crucial role in the subsequent \textit{denoising} process. To account for this, we construct the noise estimator $\hat{\bm{\epsilon}}(\cdot)$ using a residual neural network, as illustrated in Fig.~\ref{fig:noise estimator flowchart}, and train it by minimizing the following loss function:
\begin{equation}
    \bm{\mathcal{L}}_{DDPM} = {\mathbb{E}} [\left\| \hat{\bm{\epsilon}}_{\tau}(\bm{o}, \bm{a}_{\tau}, {\tau}) - {\bm{\epsilon}}_{\tau}  \right\|_{2}^{2}] , \label{eq:diffusion_loss}
\end{equation}
where $\bm{o}$ includes both the current and previous observations, as incorporating historical information helps identify trends and enhances the accuracy of predicting future actions. The diffusion step $\tau$ serves as positional information, enabling the network to recognize the current diffusion stage effectively~\cite{li2023transformer}. 

\subsubsection{Denoising Process\label{sec:DDPM_denoising}}
In contrast to the diffusion process, the \textit{denoising} process reconstructs data from noise in reverse, illustrated by the linen block in Fig.~\ref{fig:IL_overview_new}. Leveraging the previously trained noise estimator $\hat{\bm{\epsilon}}(\cdot)$, the model progressively removes the noise from a random sample $\bm{a}_{T} \in \mathcal{N}(\bm{0}, \bm{I})$, following the denoising rule:
\begin{align}   
    {\sigma}_{\tau} = & \sqrt{{{\beta}_{\tau}}} \label{eq:denoising_sigma_beta},  \\
    \Bar{{{\alpha}}}_{\tau} = & \prod_{i=1}^{\tau} {{{\alpha}}}_{i} \label{eq:denoising_alphabar_alpha},  \\
    \bm{a}_{\tau - 1} = & \frac{1}{\sqrt{\alpha_{\tau}}} \ [\bm{a}_{\tau} - \frac{1 - \alpha_{\tau}}{\sqrt{1-\Bar{\alpha}_{\tau}}} \ \hat{\bm{\epsilon}}_{\tau}( \bm{o}, \bm{a}_{\tau}, {\tau})] + \sigma_{\tau} \ \bm{\epsilon}_{\tau},  \label{eq:denoising_law}
\end{align}
where the variance schedule parameters $\Bar{{{\alpha}}}_{\tau}$ and ${\sigma}_{\tau}$ modulate the subtracted noise in each step. After $T$ steps (diffusion horizon) iteration, we obtain a probabilistic reconstructed action $\bm{a}_{0}$. An illustrative example is provided in Sec.~\ref{validation}.

\subsection{Impedance Control with Feed-forward Force \label{sec:controller}}
Consider a torque-controlled robot with $n$-Degree of Freedom, the second-order rigid body dynamics is written as:
% \mathrm{}
\begin{equation}
    \bm{M}(\bm{q})\ddot{\bm{q}}+\bm{C}(\bm{q},\dot{\bm{q}})\dot{\bq} + \bm{g} (\bq ) = \bm{\tau}_m + \bm{\tau}_{ext}, 
	\label{equ:RobotDynamics}
\end{equation}
where $\bq \in \mathbb{R}^{n}$ is the joint state. $\bm{M}(\bm{q}) \in \mathbb{R}^{n \times n}$ corresponds to the mass matrix, $\bm{C}(\bm{q},\dot{\bm{q}}) \in \mathbb{R}^{n \times n}$ is the Coriolis matrix and $\bm{g}(\bm{q}) \in \mathbb{R}^{n}$ is the gravity vector. The motor torque (control input) and external torque are denoted by $\bm{\tau}_m \in \mathbb{R}^{n}$ and $\bm{\tau}_{ext} \in \mathbb{R}^{n}$, respectively. The impedance control law with feed-forward force profile is defined as~\cite{yang2011human}:
\begin{equation}
\begin{aligned}
    \btau_m(t) = & \bm{J}( \bq)^\mathsf{T} [ \bm{F}_{ff}(t) + \bm{K}(t)\bm{e} + \bm{D} \dot{\bm{e}}  \\
                 & + \bm{M}(\bm{q})\ddot{\bx}_d] + \bm{C}(\bm{q},{\dot{\bq}})\dot{\bq} + \bm{g}(\bm{q}) ,\label{eq:control_law} %+ \btau_r\text{,}
\end{aligned}
\end{equation}
where $\bm{F}_{ff}(t)$ donates the feed-forward wrench, $\bx_d$ is desired trajectory. $\bm{x}$ indicates the robot's current position. $\bm{e}=\bm{x}_d - \bm{x}$ and $\dot{\be}={\dot{\bx}}_d - \dot{\bx}$ are the position and velocity error, respectively. $\bm{K}(t)$ and $\bm{D}$ are stiffness and damping matrices in Cartesian space. $\bm{J}(\bm{q})$ represents the robot Jacobian matrix. The internal wrench $\bm{F}_{in}$ applied by the robot on objects is calculated with: 
\begin{align}
    \bm{J}_{binv} &= \bm{J}_{body}^{\dag}, \label{eq: pseudo-inverse}  \\
    \bm{F}_{in} &= \bm{J}_{binv}^{\mathsf{T}} (\bm{\tau}_m - \bm{C}\left(\bm{q},\dot{\bq}\right)\dot{\bq} - \bm{g}\left(\bm{q}\right) ),  \label{eq: external_wrt_EE} 
\end{align}
where $\bm{J}_{binv}$ represents the pseudo-inverse of the body Jacobian $\bm{J}_{body}$, which relates joint velocities to the End-Effector (EE) twist expressed in the body frame (a frame at the EE).

\subsection{Dynamic System based Filter \label{sec: filter}}
To solve the frequency misalignment between the diffusion model and the impedance controller with feed-forward force, we interpolate the diffusion model's output $\bm{F}_{df}$ with a dynamic system-based filter, according to the equation:
\begin{equation}
    \ddot{\bm{F}}_{ff} = \alpha(\beta(\bm{F}_{df} - \bm{F}_{ff}) - \dot{\bm{F}}_{ff}), \label{eq: ds-filter}
\end{equation}
where the $\bm{F}_{df}$ refers to the raw output of the diffusion model and $\bm{F}_{ff}$ indicates the filtered and interpolated $1000$~Hz feed-forward force to be executed by the controller. The derivative and second-order derivative of $\bm{F}_{ff}$ are initialized as zero vectors. $\alpha$ and $\beta$ are two constant scales.\footnote{In this work, $\alpha$ and $\beta$ are fixed as $0.9$ and $0.3$, respectively, based on several trials that demonstrated their effectiveness.}

%% file: 3_experiment.tex
\section{Experiment}
To evaluate our proposed method, we designed a set of experiments to: \il{\item demonstrate the effectiveness of our proposed framework and validate its capability to generalize to novel tasks, \item provide a practical guideline for balancing inference ability and speed by evaluating the performance of models with varying sizes, and \item showcase the feasibility of our designed dynamic system-based filter to mitigate the frequency misalignment between diffusion model and real-time controller.}
\begin{figure}[h]
    \centering
    \includegraphics[width=0.85\linewidth]{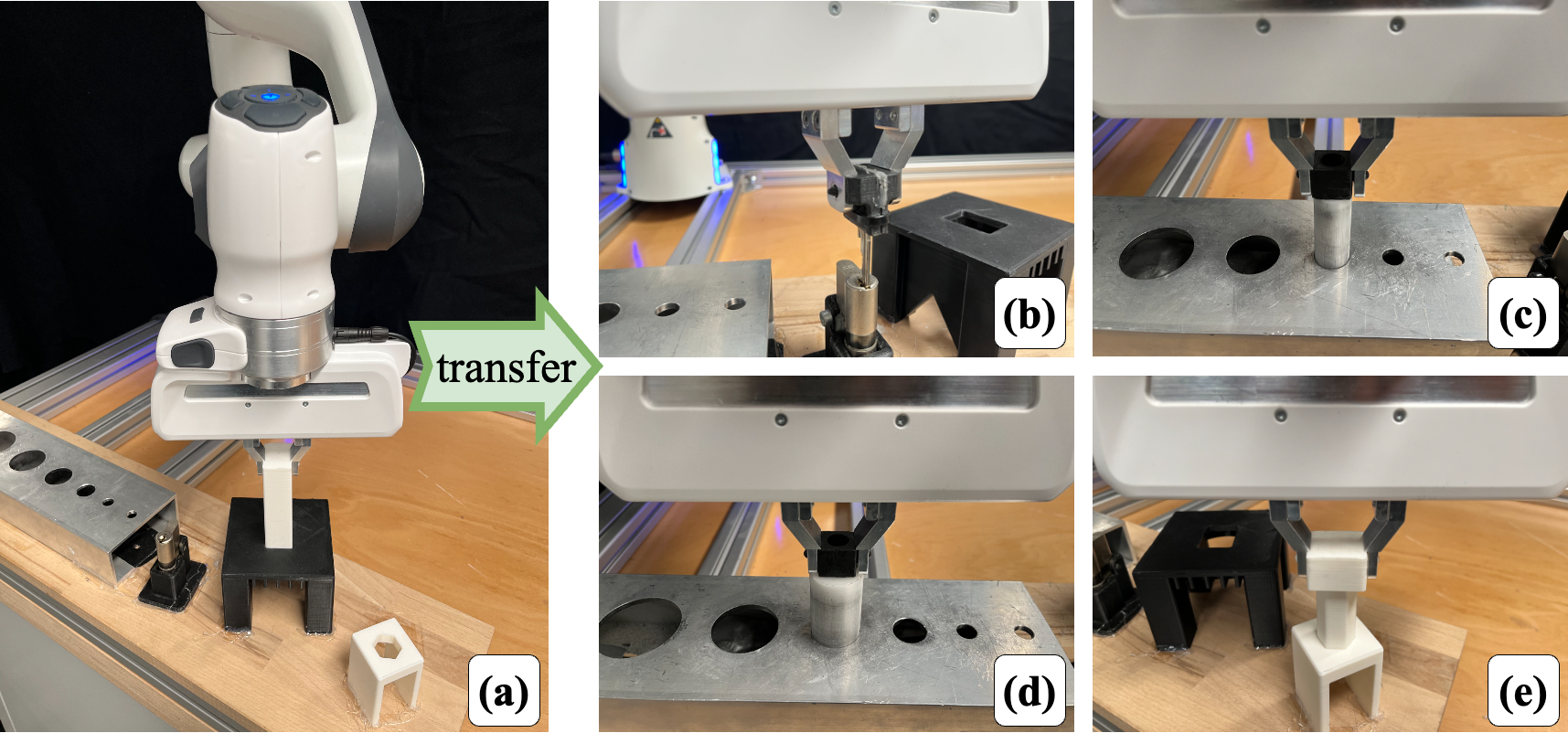}
    \caption{Experiment Setup. The object grasped by the robot in the left figure is the training object: (a) \textbf{Cuboid}: A 35~mm~×~25~mm~×~60~mm dimensional cuboid (0.1~mm clearance). The four objects on the right are applied to validate the transferability: (b) \textbf{Key}: A 37~mm long key; (c) \textbf{Cyl-S}: A 50~mm long cylinder with a diameter of 20~mm (0.02~mm clearance); (d) \textbf{Cyl-L}: A cylinder with a length of 50~mm and diameter of 30~mm (0.025~mm clearance); (e) \textbf{Prism}: A 50~mm long octagonal prism with a side length of 11 mm (0.05~mm clearance).}
    \label{fig:exps setup}
\end{figure}

\subsection{Experiment Setup}
The experiment setup shown in Fig.~\ref{fig:exps setup} consists of a Franka Emika Panda robot with 5 tight-clearance insertion objects. The robot is controlled by a PC using Ubuntu 20.04 with Intel i9-10900K CPU and real-time kernel, and the diffusion module is implemented on the PyTorch framework. Training and inference are performed on another PC with NVIDIA RTX 3090 GPU and CUDA Toolkit. 

\subsection{Data Collection \& Training \label{sec:dataset}}
% \vspace{-15pt}
\begin{figure}[h]
    \centering
    \includegraphics[width=1\linewidth]{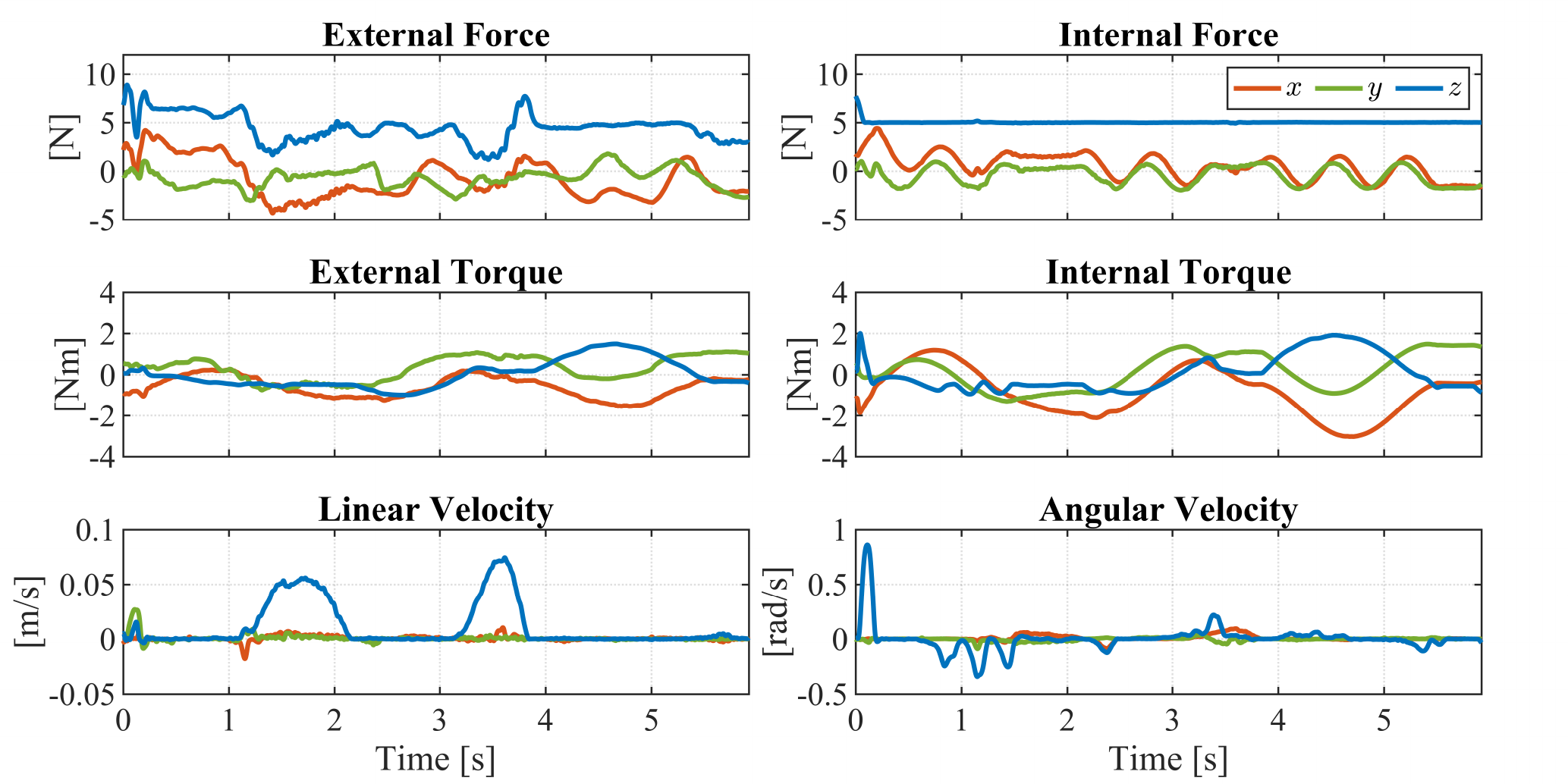}
    \caption{An example view of observations in the dataset.}
    \label{fig:collected data}
\end{figure}

\begin{figure}[htb]
    \centering
    \begin{subfigure}[b]{0.45\linewidth}
        \centering
        \includegraphics[width=\linewidth]{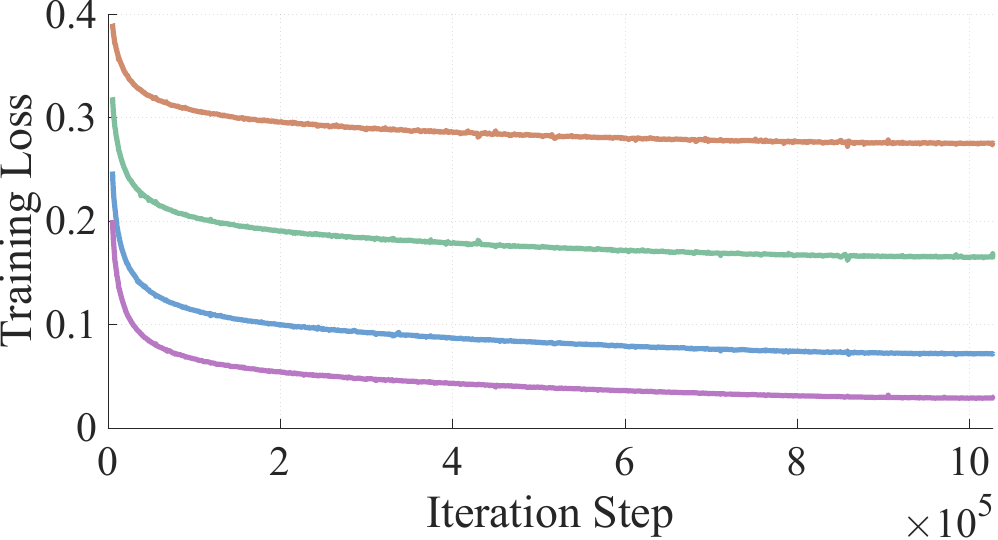}
        \caption{training loss}
        \label{fig:subfig1}
    \end{subfigure}
    \hfill
    \begin{subfigure}[b]{0.45\linewidth}
        \centering
        \includegraphics[width=\linewidth]{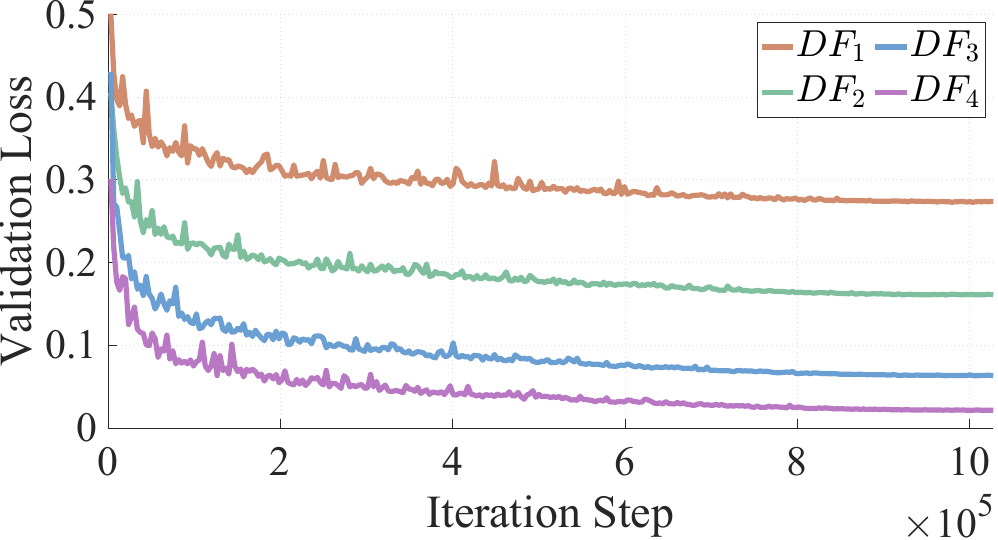}  % Replace with your second figure
        \caption{validation loss}
        \label{fig:subfig2}
    \end{subfigure}
    \vspace{7pt}
    \caption{Training loss and validation loss. Validation is conducted every 5 epochs throughout the training process.}
    \label{fig:training loss}
\end{figure}
\subsubsection{Data Collection} To train the diffusion model, we collect a comprehensive dataset comprising 1500 expert demonstrations of the assembly task, using the setup shown in Fig.~\ref{fig:exps setup}. Demonstrations are generated by executing our previous method~\cite{10610835} to perform the insertion task (Cuboid) in various initial poses. The data is recorded at 1000~Hz, resulting in a 24-dimensioned sequence, \textit{i.e.,} an 18-dimensional observation $\bm{o}$ which includes external wrench, internal wrench, and EE's speed (Fig.~\ref{fig:collected data}), paired with corresponding 6-dimensional actions $\bm{F}_{ff}$.

\vspace{5pt}
\begin{table}[H]
\centering
% \caption{Hyperparameters}
\caption{Hyperparameters for Training Diffusion Models}
\label{tab:Diffusion Model Parameter}
\begin{tabular}{ll}
\toprule
Hyperparameters & Value  \\ \hline
 Epoch &   1500    \\
 Batch Size  &   4096  \\
 Learning Rate  & $10^{-3}$         \\
 % State History Horizon ($\bm{\Delta t}$) &  7ms      \\
 Diffusion Horizon ($T$)  & 50 \\
 Diffusion Weight ($\beta_{\tau}$) &  increased from $10^{-4} \text{ to } 10^{-2}$    \\ 
 \bottomrule
\end{tabular}
\end{table}

\subsubsection{Training} 
% Although a dynamic system-based filter is designed in Sec.~\ref{sec: filter} to interpolate and smooth the diffusion model's output, a trade-off remains between the model's inference ability and speed. 
% While selecting an optimal model, there is a trade-off. 
There is a trade-off to select the optimal model. Larger models offer stronger inference capabilities, but smaller models provide faster inference speeds that are better suited to our controller. Therefore, an appropriate size is crucial for balancing performance and real-time control requirements, especially in our scenario where computational efficiency is critical.
\vspace{5pt}
\begin{table}[htb]
\centering
\caption{Details of four Diffusion Models \label{table: DFs details}}
\label{tab:models}
\begin{tabular}{c|c c c}
\toprule
Model & Neurons (N) & Final Loss & Inference Frequency \\ 
\midrule
$DF_{1}$       & $128$     &   $0.2751$   &      $503.8$~Hz         \\ 
$DF_{2}$       & $256$     &   $0.1653$   &      $297.5$~Hz         \\ $DF_{3}$       & $512$     &   $0.0716$   &      $141.8$~Hz         \\
$DF_{4}$       & $1024$    &   $0.0288$   &      $51.2$~Hz         \\ \bottomrule
\end{tabular}
\end{table}

To address this problem, we train diffusion models with varying neuron numbers $N$  (highlighted in red in Fig.~\ref{fig:noise estimator flowchart}) to provide a practical guideline. 80\% of the data is used as training data. Hyperparameters employed in this process are detailed in Table~\ref{tab:Diffusion Model Parameter}. Moreover, all trained models were exported to the \textit{ONNX} format to optimize the inference speed. Table~\ref{table: DFs details} provides the details of each model. In addition, as shown by the corresponding learning curve in Fig.~\ref{fig:subfig1}, all the candidate models successfully converge within 1,000,000 iteration steps. As the model size increases, there is a clear improvement in accuracy on the training dataset, evidenced by the decreasing final loss. However, larger models also require more computational resources, leading to an evident frequency drop from $503.8$~Hz to $51.2$~Hz.

\begin{figure}[ht]
    \centering
    \includegraphics[width=0.98\linewidth]{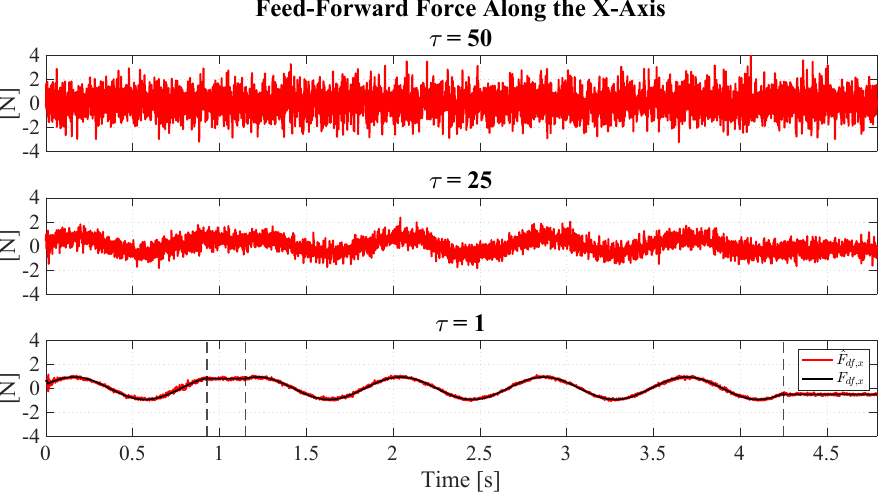}
    \caption{Denoising process with model $DF_{3}$. From the top down,  the red curves indicate the change in the diffused actions during the denoising process. The black refers to the corresponding ground truth.} \label{fig:model_validation_denoising_process_force}
\end{figure}

\subsubsection{Validation\label{validation}} The remaining 20\% of the data is used for validation. The validation losses in Fig.~\ref{fig:subfig2} imply that models have successfully converged without overfitting. Fig.~\ref{fig:model_validation_denoising_process_force} provides an intuitive instance of the denoising process, where the diffusion model reconstructs actions by progressively removing noise from a random Gaussian sample ($\tau = 50$). After 25 backward diffusion steps ($\tau = 25$), the model’s output exhibits a tendency towards the ground truth. By the final step ($\tau = 1$), the model's prediction closely matches the ground truth.

It is noteworthy that the diffusion model successfully inherits the self-adaptability of our previous method, selecting appropriate primitives based on the assembly state. The model performs a wiggle motion to align the object with the hole before $0.9$~s, and to resolve a stuck state from $1.2$~s to $4.2$~s. When the object is properly aligned, it applies a force to push the object into the insertion hole.

\subsection{Real-World Experiment Performance}
\subsubsection{Performance Test\label{sec: Feasibility test}}

% In this section, we further validate the performance of our diffusion models with the experiment setup shown in Fig.~\ref{fig:exps setup}. The object cuboid, the only object used in the dataset, is employed to assess the real-world performance of our method. Additionally, four novel objects are used to evaluate the model's transferability. For the comparison baseline, we trained a model on the cuboid assembly task using the same method employed to generate the demonstrations.

In this section, we validate the efficacy of our diffusion models using the experimental setup depicted in Fig.~\ref{fig:exps setup}. Among all demonstrated policies, we select the best-performing one as our baseline. 
We evaluate not only the performance of the candidates on the training object but also emphasize their zero-shot transferability to four novel objects.

\begin{figure}[ht]
    \centering
    \includegraphics[width=0.95\linewidth]{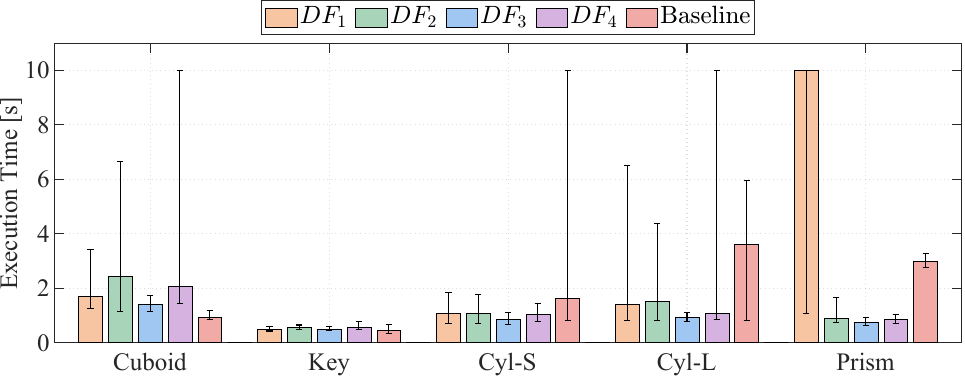}
    \caption{Execution time. The colored bars represent the median execution time for each model, and the black lines denote their 25\textsuperscript{th} and 75\textsuperscript{th} percentiles.}
    \label{fig:exps box plot}
\end{figure}

As depicted in Table~\ref{tab:model performance}, a total of 25 test cases are created by combining the models with various objects. For each case, the model is evaluated on the corresponding task with 50 random initial poses. At each pose, the robot performed two insertion trials to account for variability and reduce the influence of random occurrences. Consequently, the success rate and corresponding execution time are represented in Table~\ref{tab:model performance} and Fig.~\ref{fig:exps box plot}, respectively.

% \vspace{5pt}
% \begin{table}[H]
% \centering
%    \caption{successful rate}
% \label{tab:model performance}
% \begin{tabular}{|l|l|l|l|l|l|}
% \hline
% & Baseline & $DF_{1}$ & $DF_{2}$ & $DF_{3}$ & $DF_{4}$ \\ \hline
% Cuboid & $92.0 \ \%$ & $90.0 \ \%$& $ 79.0 \ \%$& $ \bm{98.0} \ \%$& $73.0 \ \%$\\ \hline
% Key & $94.0 \ \%$ & $ \bm{99.0} \ \%$& $94.0 \ \%$ & $ \bm{99.0} \ \%$ & $85.0 \ \%$ \\ \hline
% Cyl-S & $61.0 \ \%$ & $86.0 \ \%$& $87.0 \ \%$& $ \bm{97.0} \ \%$& $90.0 \ \%$\\ \hline
% Cyl-L & $82.0 \ \%$ & $85.0 \ \%$& $90.0 \ \%$& $ \bm{96.0} \ \%$& $66.0 \ \%$\\ \hline
% Prism & $ \bm{96.0} \ \%$ & $40.0 \ \%$& $79.0 \ \%$ & $91.0 \ \%$ & $85.0 \ \%$ \\ \hline
% Overall & $85.0 \ \%$ & $80.0 \ \%$ & $85.8 \ \%$ & $ \bm{96.2} \ \%$ & $79.8 \ \%$ \\ \hline
% \end{tabular}
% \vspace{10pt}
% \caption*{\footnotesize *The highest success rate for each task is highlighted in bold font.}
% \end{table}

\begin{table}[t]
\vspace{15pt}
\centering
   \caption{Success Rate~[\%]}
\label{tab:model performance}
\resizebox{\linewidth}{!}{
\begin{tabular}{l|l|l l l l l}
\toprule
\multirow{2}{*}{Model} & \multicolumn{1}{c|}{trained} & \multicolumn{5}{c}{novel (zero-shot transfer)} \\ \cline{2-7}
\rule{0pt}{8pt} % Adjust this value as needed
       & Cuboid & Key & Cyl-S & Cyl-L & Prism & Average \\ 
\midrule
$DF_{1}$ & $90.0$ & $ \bm{99.0} $ & $86.0 $ & $85.0 $ & $40.0 $ & $77.5 $ \\ 
$DF_{2}$ & $79.0 $ & $94.0 $ & $87.0 $ & $90.0 $ & $79.0 $ & $87.5 $ \\ 
$DF_{3}$ & $ \bm{98.0} $ & $ \bm{99.0} $ & $ \bm{97.0} $ & $ \bm{96.0} $ & $91.0 $ & $\bm{95.7} $ \\ 
$DF_{4}$ & $73.0 $ & $85.0 $ & $90.0 $ & $66.0 $ & $85.0 $ & $81.5 $ \\ 
\midrule
Baseline & $92.0 $ & $94.0 $ & $61.0 $ & $82.0 $ & $ \bm{96.0} $ & $83.3 $ \\ 
\bottomrule
   \end{tabular}}
\vspace{15pt}
\caption*{\footnotesize *The highest success rate for each task is highlighted in bold font.
The detailed configuration of models $DF_{1}$ to $DF_{4}$ is provided in Table~\ref{table: DFs details}. 
% *The overall values represent the average success rate of each model across four novel objects.
}
\end{table}

\begin{figure}[ht]
    \centering
    \includegraphics[width=0.85\linewidth]{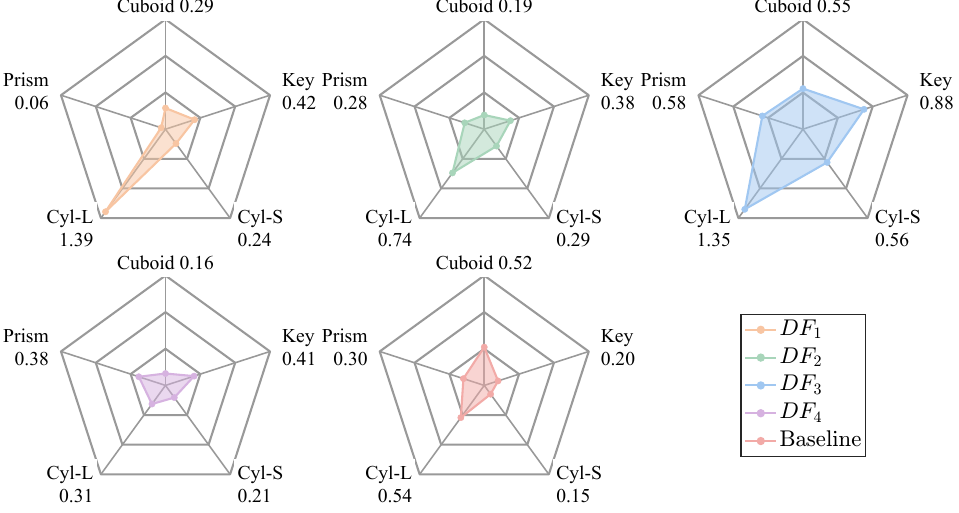}
    \caption{Radar plots for efficiency across different models}
    \label{fig: score}
\end{figure}

According to the Common Industry Format for Usability Test Reports (ISO/IEC 25062:2006), the ``core measure of efficiency” is the ratio of the task completion rate to the mean time per task~\cite{albert2013measuring}. We use this ratio as the performance metric, to evaluate the performance of comparing models. The results, illustrated by the radar plots in Fig.~\ref{fig: score}, show that $DF_3$ outperforms the baseline on demonstrated tasks in terms of efficiency.

Notably, for novel tasks, all diffusion models achieve over a 10\% improvement in efficiency, showcasing excellent zero-shot transferability. Among these models, $DF_3$ stands out with the best comprehensive performance on novel tasks, achieving an average success rate of 95.7\%.

% The results are shown as the radar plots in Fig.~\ref{fig: score}. On the demonstrated task, $DF_3$ outperforms the baseline in terms of efficiency; On novel tasks, the advantages of our diffusion model-based approach become even more pronounced. All diffusion models show significantly stronger transferability, with an efficiency improvement exceeding 10\% compared to the baseline. It is also important to emphasize that our best-performing model, $DF_3$, achieves exceptional zero-shot transferability on novel tasks, reaching an average success rate of 95.7\%. 

% As a result, DF\textsubscript{3} shows the best performance among the candidates, excelling not only on the training tasks but also on transferred tasks, leading to a 129.5\% enhancement in the overall performance compared to the baseline.

% Notably, while only DF\textsubscript{3} outperforms the baseline on the training task (Cuboid), all models based on the proposed structure demonstrate superior transferability on the novel tasks. DF\textsubscript{3} shows the best performance, excelling not only on the training tasks but also on transferred tasks, leading to a 129.5\% enhancement in the overall performance compared to the baseline.

% \begin{figure}[h]
%     \centering
%     \includegraphics[width=0.95\linewidth]{figures/score_train_novel_all.pdf}
%     \caption{score bar 5 model}
%     \label{fig: score_train_novel_all}
% \end{figure}

\subsubsection{Trade-off between model accuracy and inference speed}
As the model size increases, the model better captures latent relationships within the data, which is reflected in the increasing overall success rate from $DF_1$ to $DF_3$, as shown in Table~\ref{tab:model performance}. 
However, larger models also experience a significant reduction in inference frequency, which exacerbates the misalignment with the 1000~Hz control loop. As depicted in Table\ref{table: DFs details}, $DF_3$ maintains an acceptable frequency of 141.8~Hz, whereas $DF_4$ suffers a dramatic drop to only 51.2~Hz. This extremely low output frequency limits the model’s deployment potential despite its strong inference capability, resulting in an overall significant performance drop. Consequently, $DF_3$ (with $N=512$) is the only model that outperforms the baseline on both demonstrated and novel tasks. It exhibits the most balanced and highest performance across all insertion tasks, achieving a 129.5\% improvement in overall performance compared to the baseline.

\subsubsection{Dymanic system-based filter}
\begin{figure}[htb]
    \centering
    \includegraphics[width=0.85\linewidth]{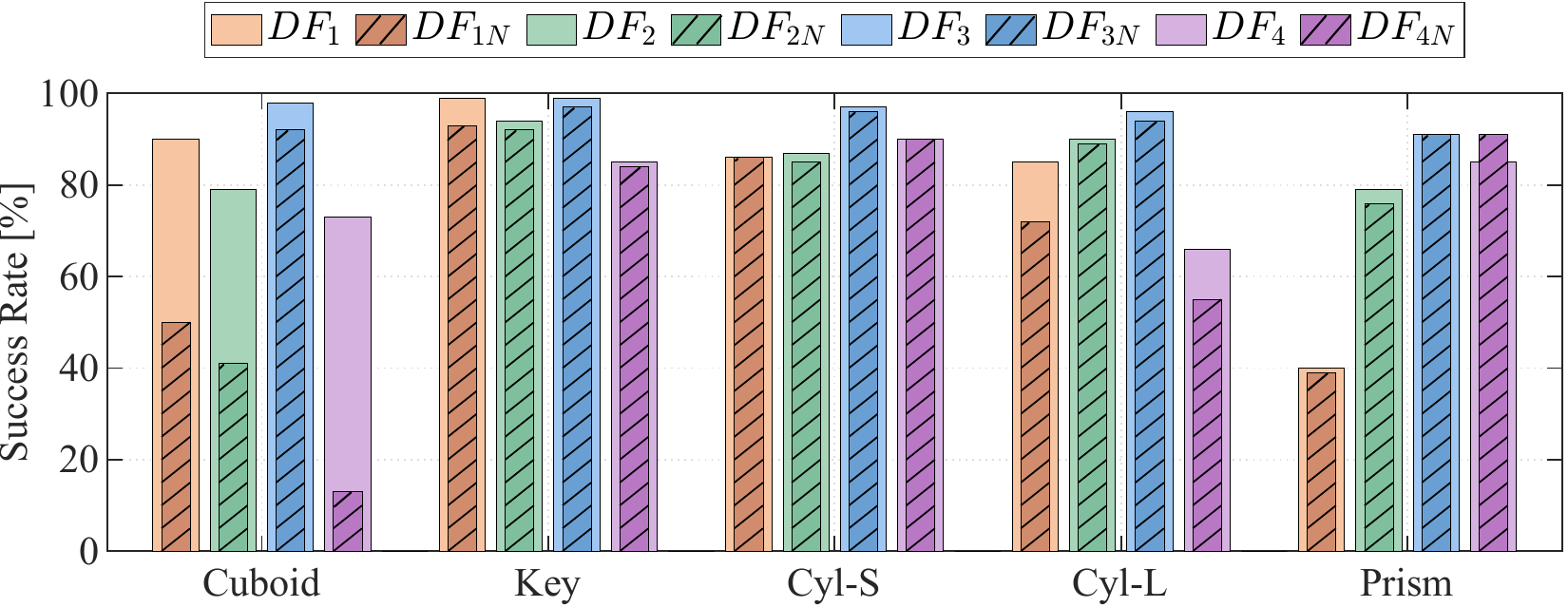}
    \caption{Impact of the dynamical system-based filter on the success rate of high-precision assembly tasks}
    \label{fig:exps box plot filter comp}
\end{figure}

% \begin{figure}[h]
%       \centering
%     \includegraphics[width=0.95\linewidth]{figures/1_cub_9_IL_ite_0_tri_1_wrench.png}
%     \caption{DF\textsubscript{3} raw output \textit{vs} executed commands with filter}
%     \label{fig:model filter figures}
% \end{figure}
Our dynamic system-based filter is designed to address the frequency misalignment issue. To validate its effectiveness, we repeat the identical experiments in Sec.~\ref{sec: Feasibility test} for the diffusion models while disabling the filter in the framework. To distinguish from the previous models ($DF_{x}$), these models are represented as $DF_{xN}$. For ease of comparison, the results are presented in the same figure. As illustrated in Fig.~\ref{fig:exps box plot filter comp}, the models with filter assistance achieve higher success rates in 16 out of 20 scenarios, with three unchanged and one decreasing by 6\%. Overall, our dynamic system-based filter mitigates the effects of frequency misalignment, leading to a 9.15\% increase in success rates.

% with diffusion models while disabling the filter in the framework. 
%  and use the notation $DF_{xN}$ to represent the model $DF_{x}$ without the filter's assistance. We repeat the experiments on  in Sec.~\ref{sec: Feasibility test} with $DF_{xN}$ models.  
% These models were re-evaluated on the 5 assembly tasks under the same conditions as in our previous experiments described in Sec.~\ref{sec: Feasibility test}. 

% As illustrated in Fig.~\ref{fig:exps box plot filter comp}, the filter improves the success rate in 16 out of 20 testing scenarios. Three scenarios maintained their previous performance, while one experienced a 6\% decrease. Overall, the filter effectively mitigates the impact of reduced inference speed, resulting in a 9.15\% increase in the overall success rate.

% Fig.~\ref{fig: ranking} visualizes the performance ranking of each model, sorted in descending order of efficiency.

\begin{figure}[htb]
      \centering
    \includegraphics[width=0.95\linewidth]{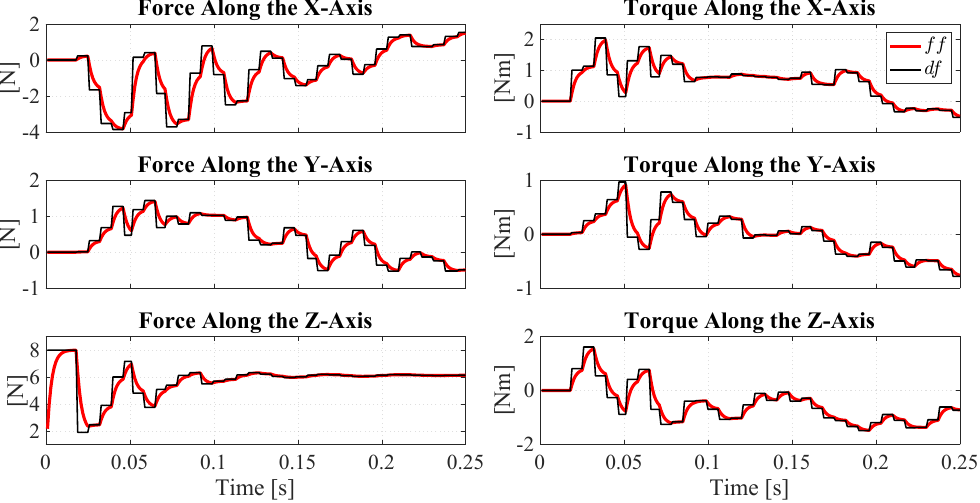}
    \caption{Impact of the filter on diffusion model's predictions. The red curves represent the filtered feed-forward wrench, while the black curves correspond to the raw outputs from the diffusion model $DF_3$.}
    \label{fig:model filter figures}
\end{figure}

Moreover, we compare the model's performance on both demonstrated and novel objects as illustrated in Fig.\ref{fig: ranking}. The inclusion of the filter results in enhanced performance across both categories. Besides, a more concrete example is provided in Fig.\ref{fig:model filter figures}, vividly illustrating the effect of our filter on diffusion model outputs. The raw diffusion output, depicted by the black curves, exhibits higher variability and fluctuations in force and torque components. In contrast, the filtered feed-forward force commands, indicated by the red curves, present a smoother profile at 1000~Hz. These results confirm that the filtering process mitigates the frequency misalignment issue.

% \begin{figure}[t]
%     \centering
%     \begin{subfigure}[b]{0.45\linewidth}
%         \centering
%         \includegraphics[width=\linewidth]{figures/score_bar_cuboid_0911_2.pdf}
%         \caption{demonstrated object}
%         \label{fig:score_subfig1}
%     \end{subfigure}
%     \hfill
%     \begin{subfigure}[b]{0.45\linewidth}
%         \centering
%         \includegraphics[width=\linewidth]{figures/score_bar_others_0911_2.pdf}  % Replace with your second figure
%         \caption{novel objects}
%         \label{fig:score_subfig2}
%     \end{subfigure}
%     \vspace{7pt}
%     \caption{Models' efficiency sorted in descending order.}
%     \label{fig: ranking}
% \end{figure}

% In terms of efficiency, the performance ranking of each model, sorted in descending order, is shown in Fig.~\ref{fig: ranking}. The figure clearly illustrates the overall performance improvement achieved by incorporating the filter, particularly for $DF_1$, where the filter enables it to surpass the baseline.

% In addition, the filter's effect on the diffusion model's predictions is illustrated in Fig.~\ref{fig:model filter figures}. The raw output of the diffusion model $DF_3$, denoted by the black curves, exhibits higher variability and fluctuations across the force and torque components. In contrast, the filtered feed-forward force commands, represented by the red curves, demonstrate a smoother profile at a frequency of 1000~Hz. These results conclude that the filtering process alleviates the negative impact of the model's lower inference speed, effectively mitigating the conflict between inference accuracy and execution speed. 

% The black curves refer to the DF\textsubscript{3}'s raw output $\bm{F}_{df}$, and the red ones indicate the corresponding filtered feed-forward force commands $\bm{F}_{ff}$.  As illustrated, the filter effectively smooths the step-shaped curve produced by the diffusion model. 

% In conclusion, by counteracting the negative effect of the slow-down model on manipulation performance, our dynamic system-based filter successfully alleviates the conflict between model accuracy and inference speed.

% \begin{figure}[h]
%     \centering
%     \includegraphics[width=0.95\linewidth]{figures/score_bar_subplots.pdf}
%     \caption{score bar 9 model, left: cuboid, right: average}
%     \label{fig:score bar}
% \end{figure}

% \begin{figure}[h]
%     \centering
%     \includegraphics[width=0.95\linewidth]{figures/score_spider_split.pdf}
%     \caption{score spider rank 5 model}
%     \label{fig:score spider}
% \end{figure}

%% file: 4_conclusion.tex
\begin{figure}[htb]
    \centering
    \begin{subfigure}[b]{0.4\linewidth}
        \centering
        \includegraphics[width=\linewidth]{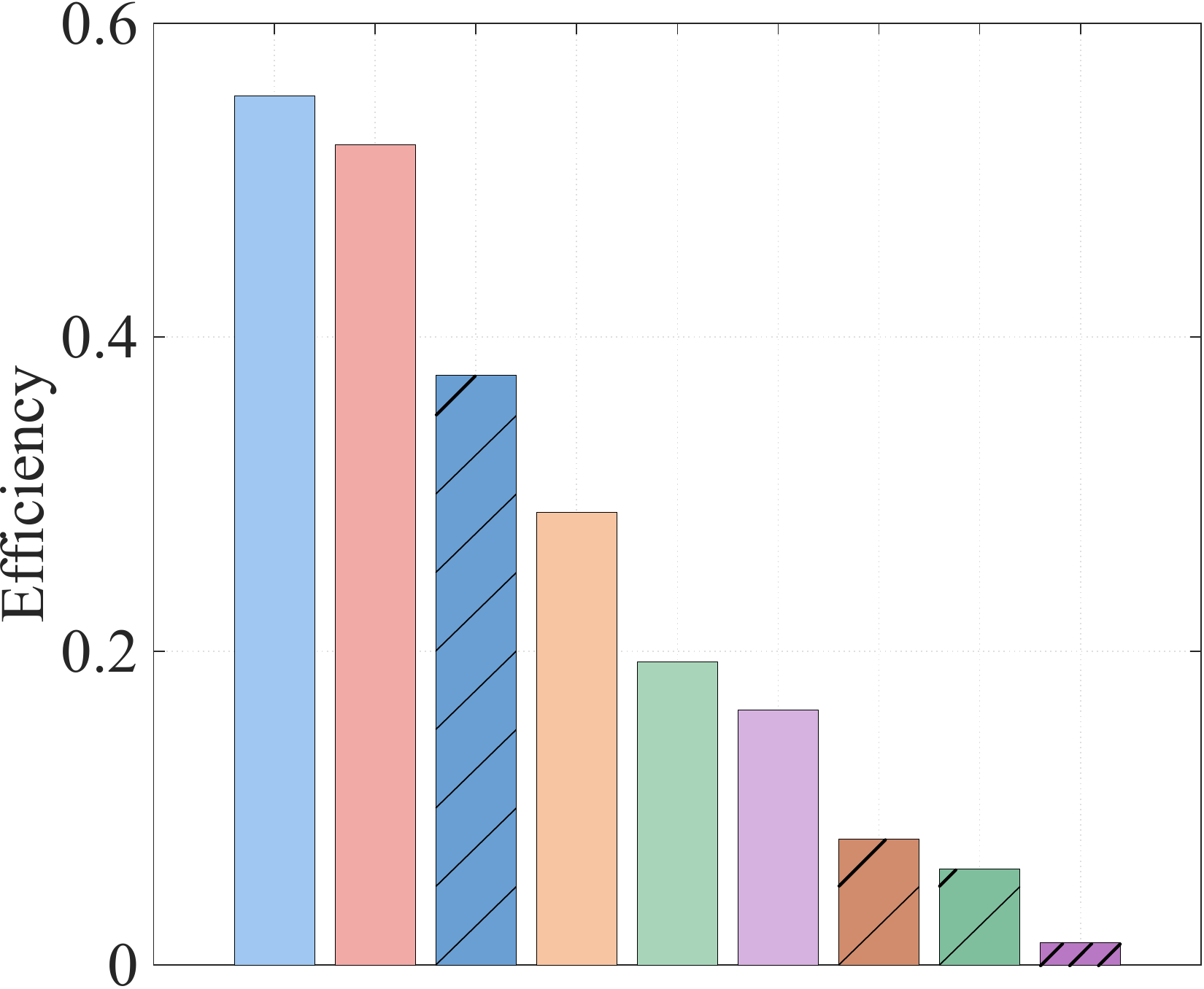}
        \caption{demonstrated object}
        \label{fig:score_subfig1}
    \end{subfigure}
    \hfill
    \begin{subfigure}[b]{0.4\linewidth}
        \centering
        \includegraphics[width=\linewidth]{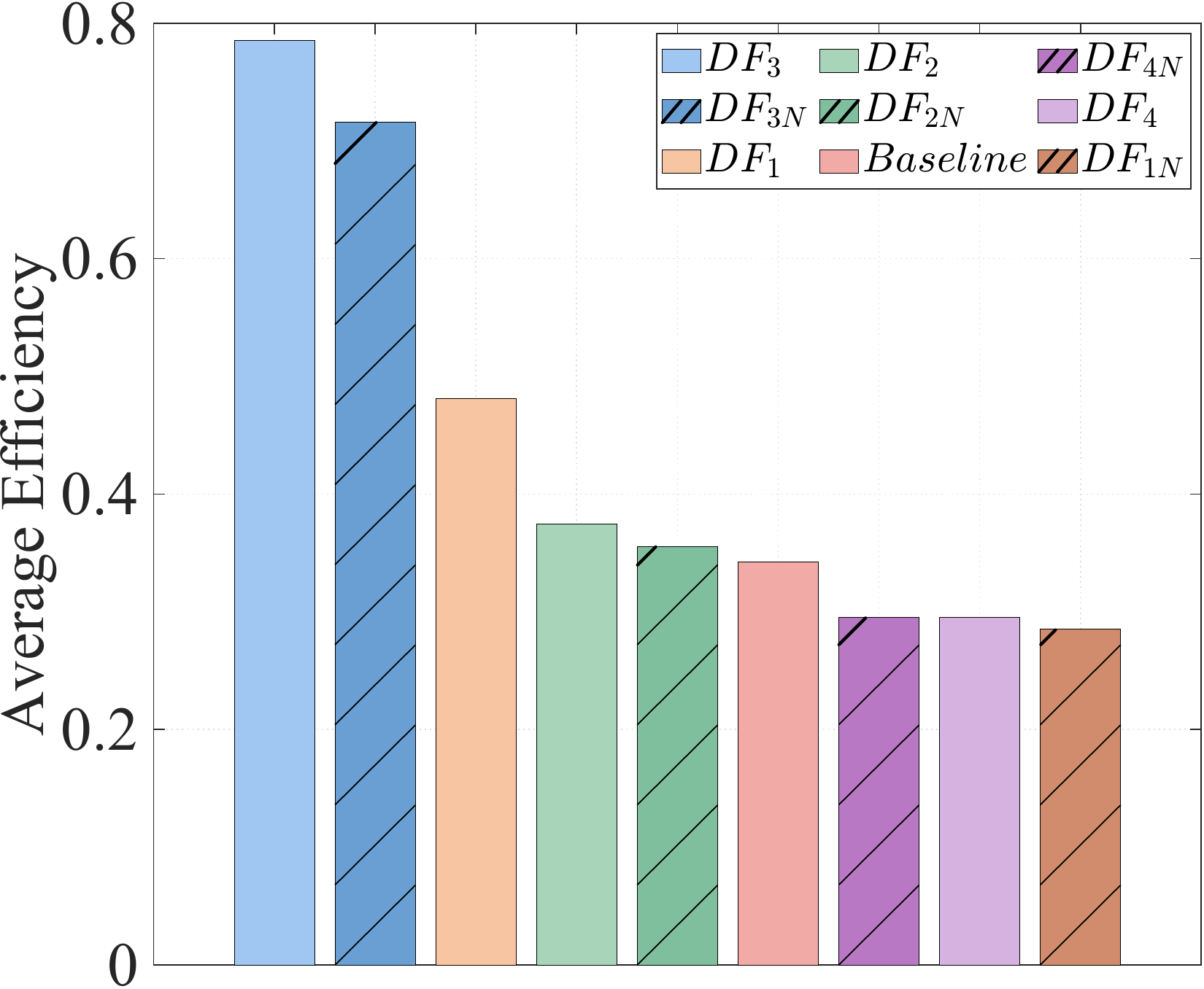}  % Replace with your second figure
        \caption{novel objects}
        \label{fig:score_subfig2}
    \end{subfigure}
    \vspace{7pt}
    \caption{Model efficiency sorted in descending order.}
    \label{fig: ranking}
\end{figure}

\section{Conclusion}
In this work, we present a novel framework leveraging diffusion models to generate 6D wrench for tactile manipulation in high-precision robotic assembly tasks. %specifically focusing on high-precision insertion tasks and the method's transferability. 
%Our approach demonstrates significant improvements in both data efficiency and task transferability compared to traditional methods.
Our approach, being the first force-domain diffusion policy, demonstrated excellent improved zero-shot transferability compared to prior work, by achieving an overall 95.7\% success rate in zero-shot transfer in experimental evaluations. Additionally, we investigate the trade-off between accuracy and inference speed and provide a practical guideline for optimal model selection.
% *In this work, we introduce an innovative framework that utilizes diffusion models to generate 6D wrench for tactile manipulation in high-precision robotic assembly tasks. As the first force-domain diffusion policy, our approach achieves an impressive 95.7\% success rate in zero-shot transfer, significantly outperforming our previous work. This improvement highlights the framework's superior zero-shot transferability. Additionally, we analyze the trade-off between accuracy and inference speed, providing practical guidelines for optimal model selection.
Further, we address the frequency misalignment between the diffusion policy and the real-time control loop with a dynamic system-based filter, significantly improving the task success rate by 9.15\%. 
%mitigates conflicts between model complexity and control frequency. 
Extensive experimental studies in our work underscore the effectiveness of our framework in real-world settings, showcasing a promising approach tackling high-precision tactile manipulation by learning diffusion-based transferable skills from expert policies containing primitive-switching logic.
% *To address the frequency misalignment between the diffusion policy and real-time control loops, we implement a dynamic system-based filter, which enhances the task success rate by 9.15\%. Our extensive experimental studies validate the effectiveness of our framework in real-world scenarios, demonstrating its potential for high-precision tactile manipulation by leveraging diffusion-based transferable skills derived from expert policies with primitive-switching logic.
%demonstrating superior performance over previous models.
% In future work, we aim to expand the applicability of this framework to more high-precision assembly tasks and explore the integration of additional sensing modalities to further improve the system’s adaptability and robustness in real-time scenarios.
In future work, we will focus on extending the framework's applicability to a broader range of high-precision assembly tasks and integrating additional sensing modalities to enhance system adaptability and robustness in real-time environments.

% There may be several possible reasons for the unsatisfactory performance of DF1 and DF4: \\
% 1. As observed in previous experiences, temperature considerably impacts model performance, possibly due to sensor noise or temperature drift. \\
% 2. The sensors may have developed defects after numerous experiments. The prism experiments are particularly complex, and the wiggling actions during object insertion are critical. These actions may have damaged the clip, as I noticed that the initial clip pose and the final insertion hole pose are more tilted than before, indicating a misalignment. \\